\documentclass{article}
\usepackage{ijcai25}

\usepackage{times}
\usepackage{soul}
\usepackage{url}
\usepackage[hidelinks]{hyperref}
\usepackage[utf8]{inputenc}
\usepackage[small]{caption}
\usepackage{graphicx}
\usepackage{amsmath}
\usepackage{amsthm}
\usepackage{booktabs}
\usepackage{algorithm}
\usepackage{algorithmic}
\usepackage[switch]{lineno}
\usepackage{amsmath}
\usepackage{amsfonts}
\usepackage{amssymb} 
\usepackage{booktabs}
\usepackage{multirow}
\usepackage{graphicx}

\title{STRGCN: Capturing Asynchronous Spatio-Temporal Dependencies for Irregular Multivariate Time Series Forecasting }

\author{
Yulong Wang
\and
Xiaofeng Hu \and
Xiaojian Cui \And
Kai Wang\thanks{Corresponding author}\\
\affiliations
College of Computer Science, Nankai University\\
\emails
yl.wang@mail.nankai.edu.cn,
\{13230958736, cuixiaojian0000\}@163.com,
wangk@nankai.edu.cn
}

\begin{document}

\maketitle

\begin{abstract}
Irregular multivariate time series (IMTS) are prevalent in real-world applications across many fields, where varying sensor frequencies and asynchronous measurements pose significant modeling challenges. Existing solutions often rely on a pre-alignment strategy to normalize data, which can distort intrinsic patterns and escalate computational and memory demands. Addressing these limitations, we introduce STRGCN, a Spatio-Temporal Relational Graph Convolutional Network that avoids pre-alignment and directly captures the complex interdependencies in IMTS by representing them as a fully connected graph. Each observation is represented as a node, allowing the model to effectively handle misaligned timestamps by mapping all inter-node relationships, thus faithfully preserving the asynchronous nature of the data. Moreover, we enhance this model with a hierarchical ``Sandwich'' structure that strategically aggregates nodes to optimize graph embeddings, reducing computational overhead while maintaining detailed local and global context. Extensive experiments on four public datasets demonstrate that STRGCN achieves state-of-the-art accuracy, competitive memory usage and training speed. 
\end{abstract}

\section{Introduction}

\begin{figure}[t!]
\centering
\includegraphics[width=1\columnwidth]{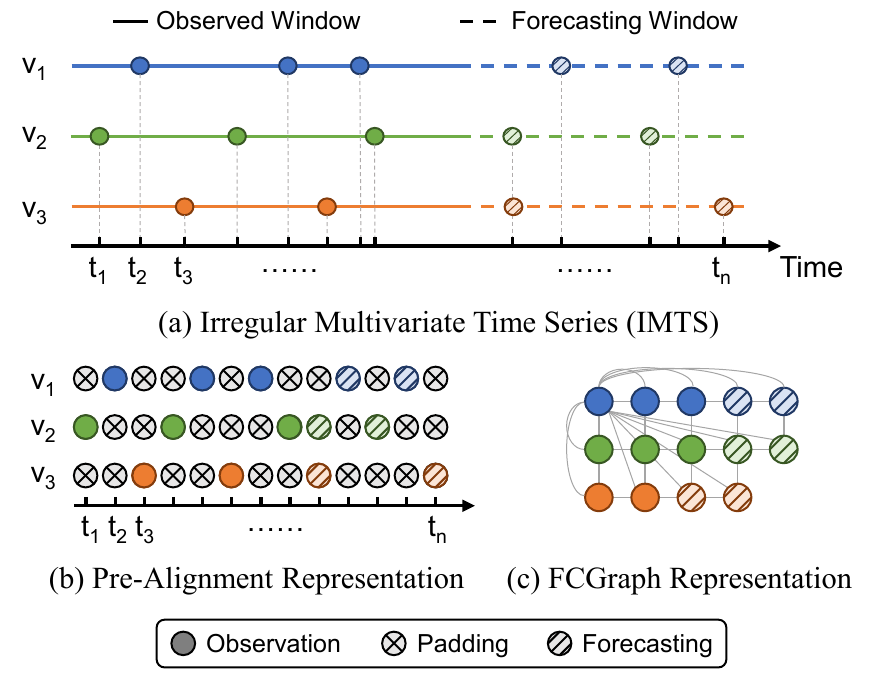}
\caption{(a) Illustrates the characteristics of an irregular multivariate time series, where variable \(v_1\) has irregular sampling intervals, and \(v_1\), \(v_2\), and \(v_3\) have different sampling frequencies and misaligned timestamps.  
(b) displays the pre-alignment representation for IMTS, which results in significant additional memory consumption due to the inclusion of padded values.  
(c) shows the fully-connected graph representation for IMTS, where each observation is treated as a node, and some edges are omitted for clarity.  }

\label{fig1}
\end{figure}

In recent years, multivariate time series forecasting (MTSF) has seen significant advancements, with numerous models and methods being developed to address various challenges \cite{TFB}. However, most of these studies share a common assumption: time series are sampled at regular intervals, and observations across variables are temporally aligned. This assumption often diverges from real-world scenarios \cite{weerakody2021review}. For instance, in industrial applications, sensors may operate at varying sampling frequencies due to design requirements or hardware limitations \cite{steed2017falcon,yuan2021sampling}. Similarly, in healthcare, different measurements are taken at irregular intervals due to testing procedures, patient conditions, or other practical constraints \cite{johnson2016mimic,zhang2023warpformer}. These characteristics, as illustrated in Figure \ref{fig1}a, render traditional methods for forecasting regularly sampled multivariate time series inapplicable to irregular multivariate time series (IMTS) \cite{zhang2023warpformer}.

To address the challenges of IMTS forecasting, researchers have proposed various approaches. 
A straightforward method involves using interpolation or probabilistic techniques to transform irregularly sampled data into a regularly multivariate time series, which are then processed using common models \cite{schafer2002missing}. 
However, existing work has shown that these methods can lose critical information about missing patterns during the interpolation stage \cite{zhang2022graphguided}. 
Furthermore, efforts have been made to handle irregular sampling in univariate or low-dimensional series using neural ordinary differential equations (Neural ODEs) \cite{chen2018neural,rubanova2019latent}, but these methods still face limitations in managing asynchronous observations and inter-variable dependencies \cite{zhang2024irregular}.

Recent deep learning models have attempted to directly model IMTS by explicitly incorporating timestamps or sampling intervals into their input features\cite{shukla2021multitime,tipirneni2022self,zhang2022graphguided,zhang2023warpformer}. 
However, due to the stringent requirements on input size and format imposed by deep networks, existing methods typically employ a "Pre-Alignment Representation" strategy \cite{che2018recurrent}, as shown in Figure \ref{fig1}b. This strategy involves extending each univariate observation to include all observed timestamps. Consequently, the input matrix size grows exponentially with the number of variables, leading to excessive memory usage and computational costs during both training and inference \cite{zhang2024irregular}.

Against this backdrop, we propose a novel Spatio-Temporal Relational Graph Convolutional Network (STRGCN) model, designed to fundamentally resolve these challenges. The model addresses two key challenges:

\textbf{Adaptability to Irregular Multivariate Time Series:}
We represent the entire IMTS as a fully connected graph (as depicted in Figure \ref{fig1}c), where nodes correspond to observation points, and edges represent potential relationships between any two points. Compared to traditional "temporal pre-alignment" methods, this graph representation effectively prevents exponential growth in the size of the input matrix. It enables the model to naturally process IMTS data with varying sampling intervals and misaligned timestamps without the need to map them onto a regular temporal axis.

\textbf{Modeling Asynchronous Spatio-temporal Dependencies:}
 With the IMTS represented as a fully connected graph, we introduce the STRGCN, which integrates ideas from Relational Graph Convolutional Networks (RGCNs) \cite{schlichtkrull2018modeling} to process asynchronous observations and their interdependencies. Specifically, STRGCN employs learnable variable encodings and time encodings on graph edges, allowing it to differentiate relationships between observations from various variables and at distinct temporal distances. This design effectively captures the intricate spatio-temporal dependencies inherent in IMTS.

In addition, STRGCN incorporates an optional hierarchical ``Sandwich'' structure within its network. Observations at the base layer are selectively aggregated and mapped to an intermediate layer, where richer contextual semantics are captured before being upsampled back to the full node set. This process compensates for the sparse representation of aligned observations at the base layer while enabling the intermediate layer to capture both local and global semantic information. Furthermore, the reduced resolution of the intermediate layer decreases computational overhead when operating on the fully connected graph.

In summary, our main contributions are as follows:

\begin{itemize}
\item We propose the STRGCN model, a novel approach for forecasting IMTS that naturally adapts to irregular sampling without relying on temporal pre-alignment. By representing IMTS as a fully connected graph, the model can directly capture spatiotemporal dependencies among all observations without incurring additional memory or computational costs.
\item We introduce a hierarchical ``Sandwich'' design within the STRGCN to integrate local and global semantic relationships while reducing the computational complexity of graph convolutions on large-scale fully connected graphs.
\item Our extensive experiments on four public irregular time series datasets, compared with several state-of-the-art baseline models, demonstrate that STRGCN displays the best overall performance among existing methods in terms of prediction accuracy, memory usage, and training speed.
\end{itemize}

\section{Related Work}
\subsection{Irregular Multivariate Time Series Forecasting}

Modeling irregular multivariate time series (IMTS) is challenging due to irregular sampling intervals and asynchrony among variables. Early methods relied on interpolation \cite{schafer2002missing,xu2018raim}, often losing crucial information from the original sampling. Later, some researchers used neural ordinary differential equations (Neural ODEs) \cite{chen2018neural,kidger2020neural,schirmer2022modeling} and enhanced recurrent neural networks \cite{che2018recurrent,li2019vs,weerakody2023cyclic} to directly model on irregular intervals. While these models handled irregular sampling effectively, they struggled to capture dynamic correlations among asynchronous variables. Recent advancements have seen the adoption of Transformer \cite{tipirneni2022self,wei2023compatible,zhang2023warpformer} and graph neural networks \cite{zhang2022graphguided} to model dependencies within series and between variables. Simultaneously, these models commonly adopt a pre-alignment representation for IMTS, aligning all observations to a uniform timestamp sequence. This approach increases memory and computational costs in environments with highly asynchronous variables. Notably, the t-PatchGNN \cite{zhang2024irregular} addresses this issue by segmenting pre-aligned data into patches at fixed intervals, and adapting to different sequence lengths using an additional method similar to meta-learning. However, this approach models spatiotemporal relationships at a coarse granularity, losing node-level information. In contrast, our STRGCN leverages the fully-connected graph representation of IMTS to naturally capture spatiotemporal correlations at the node level without additional modules, efficiently adapting to asynchronous observations.

\begin{figure*}[t!]
\centering
\includegraphics[width=0.98\textwidth]{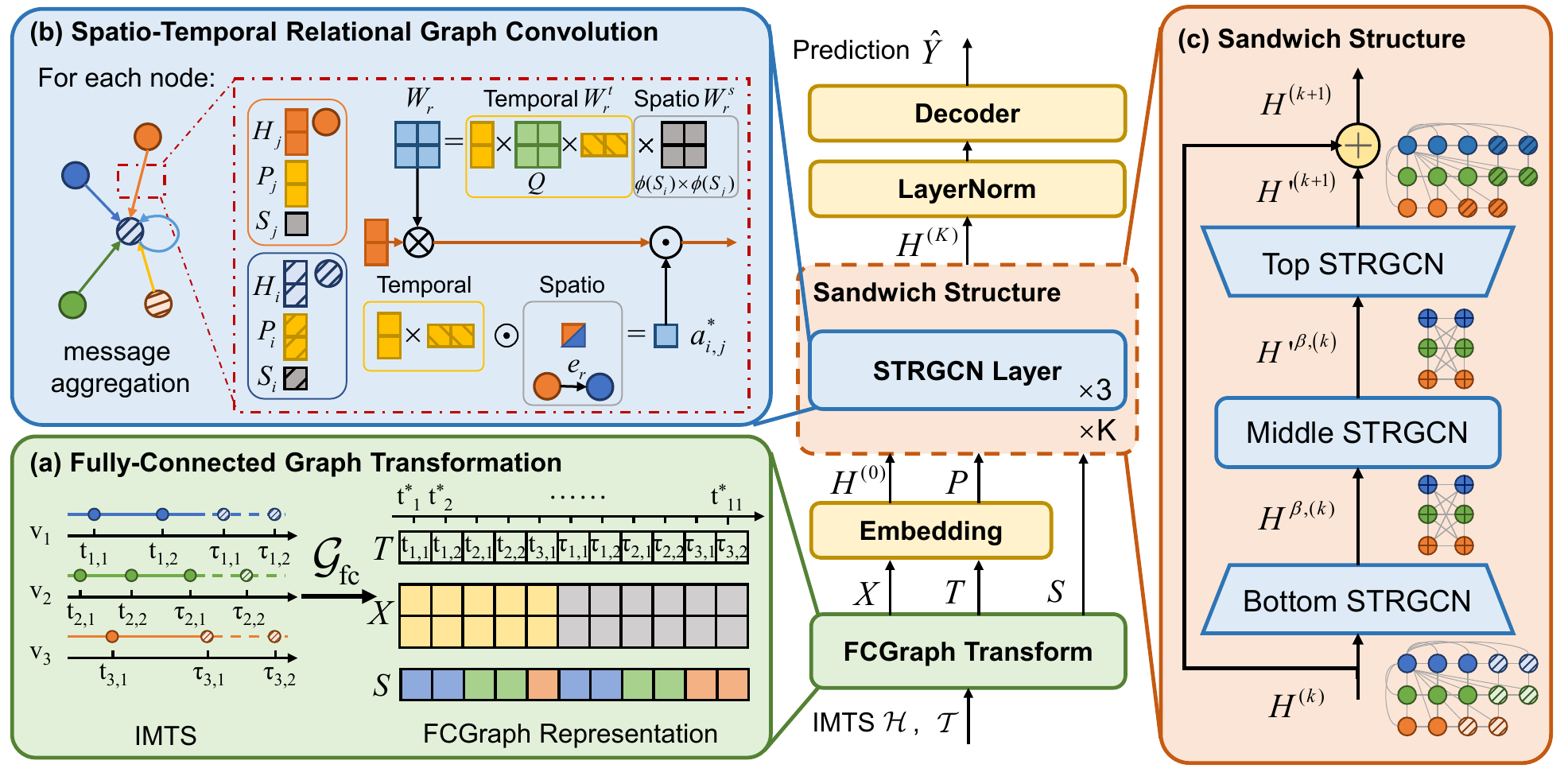} 
\caption{Overview of STRGCN, consisting of the following key components: (a) the Fully-Connected Graph Transformation Module, which converts IMTS data into a compact representation of a fully connected graph; (b) the Spatio-Temporal Relational Graph Convolution Layer, designed to capture asynchronous spatio-temporal dependencies; and (c) the Hierarchical Sandwich Structure, which integrates local and global semantic relationships while mitigating computational complexity.}

\label{main}
\end{figure*}

\subsection{Graph Based Spatio-Temporal Modeling}

Graph neural networks (GNNs) have been widely applied in spatio-temporal forecasting tasks for multivariate time series \cite{jin2024survey}. Most approaches typically use GNNs to capture spatial dependencies between variables, coupled with specialized temporal prediction networks such as TCNs to capture temporal dynamics \cite{yu2018spatio,guo2019attention,liu2022multivariate,cai2024msgnet}.
This decoupled framework effectively leverages the strong spatial modeling capabilities of GNNs.
A minority of models attempt to learn spatio-temporal dependencies using pure graph models, avoiding reliance on separate temporal networks and thereby integrating spatial and temporal information within a unified framework \cite{yi2024fouriergnn,wang2024fully}.
These methods perform well on regular multivariate time series but struggle to naturally model asynchronous spatio-temporal relationships when the sampling timestamps of different variables are misaligned.
An exception is RainDrop \cite{zhang2022graphguided}, which addresses the asynchrony of IMTS by passing messages at all timestamps where variable observations occur. However, this approach significantly increases computational costs due to the need to propagate messages across all misaligned time points.

\section{Problem Definition}

Let $v_1, \dots, v_N$ represent $N$ distinct variables. For each variable $v_n$, the set of historical observations is given by 
\(
\mathcal{H}_n = \{(t_{n,i}, z_{n,i})\}_{i=1}^{K_n} 
\) 
where $t_{n,i}$ is the timestamp and $z_{n,i}$ is the corresponding value for the $i$-th observation, and $K_n$ denotes the number of observations for variable $v_n$. The aggregate historical data across all variables is 
\(
\mathcal{H} = \{\mathcal{H}_n\}_{n=1}^N.
\)

For forecasting, for each variable $v_n$, define a set of future timestamps 
\(
\mathcal{T}_n = \{\tau_{n,j}\}_{i=1}^{M_n},
\) 
where each $\tau_{n,j} > t_{n, K_n}$. The set of all future timestamps is represented as 
\(
\mathcal{T} = \{\mathcal{T}_n\}_{n=1}^N.
\)

The task of Irregular Multivariate Time Series Forecasting involves learning a mapping function 
\begin{equation}
F : (\mathcal{H}, \mathcal{T}) \mapsto \widehat{\mathcal{Y}} = \{[\widehat{z}_{n,j}]_{j=1}^{M_n} \}_{n=1}^N
\end{equation}
where \(\widehat{\mathcal{Y}}\) contains the predicted values \(\widehat{z}_{n,j}\) for each future timestamp $\tau_{n,j}$ associated with variable $v_n$. The challenge lies in accurately predicting these future values using the irregular and non-aligned historical observations in $\mathcal{H}$ without assuming uniform sampling across variables.

\section{Method}
\subsection{Overall Architecture}

Figure \ref{main} illustrates the overarching structure of the STRGCN framework. Given an input irregular multivariate time series (IMTS) \(\mathcal{H}\) and a set of future timestamps \(\mathcal{T}\) for forecasting, we first transform these inputs into a fully connected graph representation. Specifically, we define a transformation 
\begin{equation}
(X,\, T,\, S) \;=\; \mathcal{G}_{\mathrm{fc}}(\mathcal{H},\,\mathcal{T})
\end{equation}
where \(X \in \mathbb{R}^{L \times C}\) denotes the matrix of node feature values, \(T \in \mathbb{R}^L\) represents the corresponding timestamps for each node, and \(S \in \mathbb{R}^L\) encodes the variable identifiers associated with each node. Here, the total number of nodes \(L\) equals the sum of all historical and future observations across variables.

Following the graph construction, the raw node values and timestamps are projected into latent spaces. Specifically, we compute:
\begin{equation}
H^{(0)} \;=\; \mathrm{Linear}(X), \qquad P \;=\; \mathrm{TimeEmbed}(T)
\end{equation}
where \(H^{(0)} \in \mathbb{R}^{L \times D}\) are the initial node value embeddings, and \(P \in \mathbb{R}^{L \times D}\) are the corresponding time embeddings derived from \(T\), with \(\mathrm{TimeEmbed}\) being the continuous version of the transformer positional encoder. Here, \(D\) denotes the dimensionality of the hidden layer. 

These embeddings are propagated through a multi-layer STRGCN for hierarchical feature aggregation. In this architecture, each layer, denoted by k, can represent an individual STRGCN layer or a hierarchical ``Sandwich'' structure comprising several STRGCN layers. Each layer updates the node representations according to the spatio-temporal relations as follows:
\begin{equation}
H^{(k+1)} = \mathrm{STRGCN}^{(k)}(H^{(k)},\, P,\, S,\, \Theta^{(k)})
\end{equation}
where \(\Theta^{(k)}\) represents the set of parameters specific to the \(k\)-th layer of the STRGCN model. The ``Sandwich'' design of the architecture selectively aggregates node features and subsequently upsamples them, effectively capturing both local and global spatio-temporal dependencies in a hierarchical manner.

Finally, after the last STRGCN layer, we extract the embeddings corresponding to nodes at future timestamps. Let \(\widehat{H}\) denote these embeddings. The final prediction is then obtained by passing \(\widehat{H}\) through a MLP decoder:
\begin{equation}
\widehat{Y} \;=\; \mathrm{Decoder}(\widehat{H})
\end{equation}
where \(\widehat{Y}\) contains the forecasted values for the future time points.



\subsection{Fully Connected Graph Representation of IMTS}
Given an input irregular multivariate time series (IMTS) \(\mathcal{H}\) and a set of future timestamps \(\mathcal{T}\), these inputs are first transformed into a fully connected graph representation.

We construct a fully connected graph \(\mathcal{G} = (\mathcal{N}, \mathcal{E})\), where \(\mathcal{N}\) is the set of nodes and \(\mathcal{E}\) is the set of edges between nodes. Each node corresponds to either a historical observation or a future query. All nodes are fully connected, meaning there is a directed edge between every pair.

Each node has three attributes: timestamp, feature, and variable identifier. Specifically, for each group of nodes corresponding to a variable \(n\), the following definitions apply:

\textbf{Timestamp \(t_n^*\)}: The timestamp associated with the nodes belonging to variable \(n\) is the union of historical timestamps \(t_n\) from \(\mathcal{H}_n\) and future timestamps \(\tau_n\) from \(\mathcal{T}_n\). Therefore, the length of the timestamp sequence for each node in this group is \(L_n = K_n + M_n\), where \(K_n\) and \(M_n\) are the number of historical observations and future queries, respectively:
\begin{equation}
    t_n^* = \{t_{n,i}\}_{i=1}^{K_n} \cup \{\tau_{n,j}\}_{j=1}^{M_n}
\end{equation}

\textbf{Feature \(x_n^*\)}: The feature associated with each node in this group is defined as
\begin{equation}
    x_{n,l}^* =
    \begin{cases}
    z_{n,l} & \text{if } t_{n,l}^* \in \mathcal{H}_n, \\
    z_{\text{unk}} & \text{if } t_{n,l}^* \in \mathcal{T}_n
    \end{cases}
    \quad \text{for each } t_{n,l}^* \in t_n^*
\end{equation}
where \(z_{\text{unk}}\) is a placeholder for future timestamps that is later replaced with a learnable encoding.
    
\textbf{Variable ID \(s_n^*\)}: The variable identifier associated with each node in this group corresponds to the index \(n\) of the variable for the given timestamp \(t_{n,l}^*\), and is defined as
\begin{equation}
    s_{n,l}^* = n \quad \text{for each } t_{n,l}^* \in t_n^*
\end{equation}

Subsequently, the timestamps, features, and variable identifiers across all variables are aggregated. Specifically:

\begin{equation}
T = \{ t_n^* \}_{n=1}^N, 
\hspace{0.5em} X = \{[ x_{n,l}^*]_{l=1}^{L_n} \}_{n=1}^N, 
\hspace{0.5em} S = \{ [s_{n,l}^*]_{l=1}^{L_n} \}_{n=1}^N
\end{equation}

where \(N\) is the total number of variables. The final matrices \(T\), \(X\), and \(S\) represent the complete set of timestamps, features, and variable IDs, with a total length \(L = \sum_{n=1}^{N} (K_n + M_n)\).

Thus, the input \(\mathcal{H}\) and \(\mathcal{T}\) are transformed into a fully connected graph representation, denoted as:

\begin{equation}
(X,\, T,\, S) = \mathcal{G}_{\mathrm{fc}}(\mathcal{H},\, \mathcal{T})
\end{equation}
where \(X\) is the feature matrix, \(T\) is the timestamp matrix, and \(S\) is the variable identity matrix. 
 This approach provides a compact representation of the IMTS.

 \subsection{Spatio-Temporal Relational Graph Convolution}

Our design for STRGCN is inspired by Relational Graph Convolutional Network (RGCN) \cite{schlichtkrull2018modeling}, which are tailored for graph-structured data containing nodes with multiple types of relationships. The key idea behind RGCN is to enhance the expressive power of graph convolutional networks (GCNs) by learning separate transformation matrices for different types of relationships between nodes. 

Let \(h_i^{(k)} \in \mathbb{R}^{d}\) denote the hidden representation of node \(i\) at the \(k\)-th layer. An RGCN layer updates \(h_i^{(k)}\) via:

\begin{equation}
\label{RGCN}
h_i^{(k+1)} = \sigma \bigl( \sum_{r \in \mathcal{R}} \sum_{j \in \mathcal{N}_i^r} a_{i,j} W_r^{(k)} h_j^{(k)} + a_{i,i} W_0^{(k)} h_i^{(k)} \bigl)
\end{equation}
where \(\mathcal{R}\) denotes the set of relation types, \(\mathcal{N}_i^r\) the set of neighbors for relation type \(r\),\(W_r^{(k)} \in \mathbb{R}^{d \times d}\) is the learnable weight matrix specific to relation \(r\), \(W_0^{(k)}\) is the self-connection matrix for node \(i\) and \(a_{i,j}\) a normalization factor. Here, \(\sigma(\cdot)\) is a nonlinear activation function, e.g., ReLU.

\subsubsection{Temporal and Spatial Relation Decomposition}
In knowledge graph applications, the relation types typically represent different semantic relationships between entities. However, in the context of Irregular Multivariate Time Series (IMTS), we extend RGCN to model both temporal and spatial dependencies. Specifically, for each edge \((i,j)\), we define a temporal relation reflecting the time gap between nodes, and a spatial relation capturing the variable dependencies between nodes.

Given the continuous and broad range of timestamps in Irregular Multivariate Time Series (IMTS), it is impractical to assign a unique relation type for each possible time difference. To address this issue, STRGCN decouples the spatio-temporal relationship into two distinct components: the spatial relation and the temporal relation. The spatial relation, represented by \(\mathcal{R}^S\), encompasses all possible combinations of variable pairs, which are finite and constrained by the number of variables, \(N\). The temporal relation, in contrast, is modeled through alternative methods to accommodate its uncountable nature.

Incorporating this decomposition into the RGCN framework, we approximate the weight matrix \(W_r\) as the product of the temporal and spatial components:

\begin{equation}
W_r \;\approx\; W_r^t \times W_r^s
\label{deST}
\end{equation}
To effectively model the continuum of temporal relations, 
For the temporal component \(W_r^t\), we face the challenge that the number of distinct temporal relations is potentially infinite. we apply a learnable transformation matrix \(Q \in \mathbb{R}^{D \times D}\) to the timestamp embeddings \(P_i\) and \(P_j\) of nodes \(i\) and \(j\). The \(W_r^t\) is then computed as the matrix product:

\begin{equation}
W_r^t = P_i \times Q \times P_j^T
\end{equation}

For the spatial relation \(W_r^s\), we introduce a learnable encoding function \(\phi(\cdot)\) that projects the variable identifier of each node into a latent space. The spatial relationship between two nodes \(i\) and \(j\) is then modeled by the product of their respective variable embeddings:

\begin{equation}
W_r^{s} \; \approx \; \phi(S_i) \times \phi(S_j)
\end{equation}
where \(\phi(S_i)\) and \(\phi(S_j)\) are the embeddings for the variable identifiers of nodes \(i\) and \(j\), respectively. This low-rank decomposition significantly reduces the parameter space while still effectively modeling the complex spatial dependencies between variables.

\subsubsection{Normalization Factor}

To ensure proper scaling during message aggregation, we define a normalization factor \(a^*_{i,j}\), which incorporates both the temporal and spatial components:
\begin{equation}
\label{STRGCNEND}
a^*_{i,j} \;=\; \frac{(p_i \times p_j^T) \times e_r}{\sum_{r \in \mathcal{R}^S} \sum_{j' \in \mathcal{N}_i^r} ((p_i \times p_{j'}^T) \times e_r)}
\end{equation}
where \(p_i\) and \(p_j\) are the temporal embeddings, and \(e_r\) is a learnable coefficient associated with the variable pair in relation \(\mathcal{R}^s\). This normalization ensures that the influence of distant nodes in time is reduced, as indicated by the smaller values of \(p_i \times p_j^T\).

\begin{table*}[t!]
\centering
\renewcommand{\arraystretch}{1.2} 
\resizebox{1\textwidth}{!}{\begin{tabular}{cc|cc|cc|cc|cc}
\toprule
 \multicolumn{2}{c}{Dataset}  & \multicolumn{2}{c}{PhysioNet} & \multicolumn{2}{c}{MIMIC} & \multicolumn{2}{c}{Activity} & \multicolumn{2}{c}{USHCN} \\
\cmidrule(lr){1-2} \cmidrule(lr){3-4} \cmidrule(lr){5-6} \cmidrule(lr){7-8} \cmidrule(lr){9-10}
\multicolumn{2}{c}{Metric} & MSE\(\times10^{-3}\) & MAE\(\times10^{-2}\) & MSE\(\times10^{-2}\) & MAE\(\times10^{-2}\) & MSE\(\times10^{-3}\) & MAE\(\times10^{-2}\) & MSE\(\times10^{-1}\) & MAE\(\times10^{-1}\) \\
\midrule
\multicolumn{2}{c}{DLinear} & 41.86 ± 0.05 & 15.52 ± 0.03 & 4.90 ± 0.00 & 16.29 ± 0.05 & 4.03 ± 0.01 & 4.21 ± 0.01 & 6.21 ± 0.00 & 3.88 ± 0.02 \\
\multicolumn{2}{c}{TimesNet} & 16.48 ± 0.11 & 6.14 ± 0.03 & 5.88 ± 0.08 & 13.62 ± 0.07 & 3.12 ± 0.01 & 3.56 ± 0.02 & 5.58 ± 0.05 & 3.60 ± 0.04 \\
\multicolumn{2}{c}{PatchTST} & 12.00 ± 0.23 & 6.02 ± 0.14 & 3.78 ± 0.03 & 12.43 ± 0.10 & 4.29 ± 0.14 & 4.80 ± 0.09 & 5.75 ± 0.01 & 3.57 ± 0.02 \\
\multicolumn{2}{c}{Crossformer} & 6.66 ± 0.11 & 4.81 ± 0.11 & 2.65 ± 0.10 & 9.56 ± 0.29 & 4.29 ± 0.20 & 4.89 ± 0.17 & 5.25 ± 0.04 & 3.27 ± 0.09 \\
\multicolumn{2}{c}{Graph Wavenet} & 6.04 ± 0.28 & 4.41 ± 0.11 & 2.93 ± 0.09 & 10.50 ± 0.15 & 2.89 ± 0.03 & 3.40 ± 0.05 & 5.29 ± 0.04 & 3.16 ± 0.09 \\
\multicolumn{2}{c}{MTGNN} & 6.26 ± 0.18 & 4.46 ± 0.07 & 2.71 ± 0.23 & 9.55 ± 0.65 & 3.03 ± 0.03 & 3.53 ± 0.03 & 5.39 ± 0.05 & 3.34 ± 0.02 \\
\multicolumn{2}{c}{StemGNN} & 6.86 ± 0.28 & 4.76 ± 0.19 & 1.73 ± 0.02 & 7.71 ± 0.11 & 8.81 ± 0.37 & 6.90 ± 0.02 & 5.75 ± 0.09 & 3.40 ± 0.09 \\
\multicolumn{2}{c}{CrossGNN} & 7.22 ± 0.36 & 4.96 ± 0.12 & 2.95 ± 0.16 & 10.82 ± 0.21 & 3.03 ± 0.10 & 3.48 ± 0.08 & 5.66 ± 0.04 & 3.53 ± 0.05 \\
\multicolumn{2}{c}{FourierGNN} & 6.84 ± 0.35 & 4.65 ± 0.12 & 2.55 ± 0.03 & 10.22 ± 0.08 & 2.99 ± 0.02 & 3.42 ± 0.02 & 5.82 ± 0.06 & 3.62 ± 0.07 \\
\midrule
\multicolumn{2}{c}{GRU-D} & 5.59 ± 0.09 & 4.08 ± 0.05 & 1.76 ± 0.03 & 7.53 ± 0.09 & 2.94 ± 0.05 & 3.53 ± 0.06 & 5.54 ± 0.38 & 3.40 ± 0.28 \\
\multicolumn{2}{c}{SeFT} & 9.22 ± 0.18   & 5.40 ± 0.08 & 1.87 ± 0.01 & 7.84 ± 0.08 & 12.20 ± 0.17 & 8.43 ± 0.07 & 5.80 ± 0.19 & 3.70 ± 0.11 \\
\multicolumn{2}{c}{RainDrop} & 9.82 ± 0.08 & 5.57 ± 0.06 & 1.99 ± 0.03 & 8.27 ± 0.07 & 14.92 ± 0.14 & 9.45 ± 0.05 & 5.78 ± 0.22 & 3.67 ± 0.17 \\
\multicolumn{2}{c}{Warpformer} & 5.94 ± 0.35 & 4.21 ± 0.12 & 1.73 ± 0.04 & 7.58 ± 0.13 & 2.79 ± 0.04 & \underline{3.39 ± 0.03} & 5.25 ± 0.05 & 3.23 ± 0.05 \\
\midrule
\multicolumn{2}{c}{mTAND} & 6.23 ± 0.24 & 4.51 ± 0.17 & 1.85 ± 0.06 & 7.73 ± 0.13 & 3.22 ± 0.07 & 3.81 ± 0.07 & 5.33 ± 0.05 & 3.26 ± 0.10 \\
\multicolumn{2}{c}{Latent-ODE} & 6.05 ± 0.57 & 4.23 ± 0.26 & 1.89 ± 0.19 & 8.11 ± 0.52 & 3.34 ± 0.11 & 3.94 ± 0.12 & 5.62 ± 0.03 & 3.60 ± 0.12 \\
\multicolumn{2}{c}{CRU} & 8.56 ± 0.26 & 5.16 ± 0.09 & 1.97 ± 0.02 & 7.93 ± 0.19 & 6.97 ± 0.78 & 6.30 ± 0.47 & 6.09 ± 0.17 & 3.54 ± 0.18 \\
\multicolumn{2}{c}{Neural Flow} & 7.20 ± 0.07 & 4.67 ± 0.04 & 1.87 ± 0.05 & 8.03 ± 0.19 & 4.05 ± 0.13 & 4.46 ± 0.09 & 5.35 ± 0.05 & 3.25 ± 0.05 \\
\multicolumn{2}{c}{t-PatchGNN} & \underline{4.98 ± 0.08} & \underline{3.72 ± 0.03} & \underline{1.69 ± 0.03} & \underline{7.22 ± 0.09} & \underline{2.66 ± 0.03} & \textbf{3.15 ± 0.02} & \underline{5.00 ± 0.04} & \underline{3.08 ± 0.04} \\

\midrule
\multicolumn{2}{c}{STRGCN} & \textbf{3.98 ± 0.16} & \textbf{3.63 ± 0.07} & \textbf{1.47 ± 0.05} & \textbf{6.96 ± 0.12} &  \textbf{2.53 ± 0.18} & 3.40 ± 0.09 &  \textbf{4.92 ± 0.06} & \textbf{3.01 ± 0.05} \\
\bottomrule
\end{tabular}}
\caption{Results on irregular multivariate time series forecasting, evaluated using MSE and MAE (lower is better). Each experiment was repeated three times to compute the mean and variance. The best results are highlighted in \textbf{bold}, while the second-best results are \underline{underlined}. }
\label{Tab1}
\end{table*}

\subsubsection{ Final STRGCN Update Rule }

The final message aggregation rule of STRGCN layer, which incorporates both temporal and spatial relations as derived from equations \ref{RGCN} through \ref{STRGCNEND}, is as follows:

\begin{equation}
\label{STRGCN_F}
h_i^{(k+1)} = \sigma \bigl( \sum_{r \in \mathcal{R^S}} \sum_{j \in \mathcal{N}_i^r} 
a^*_{i,j} {W_r^t}^{(k)} {W_r^s}^{(k)} h_j^{(k)} + a_{i,i} W_0^{(k)} h_i^{(k)} \bigl)
\end{equation}

This formulation allows STRGCN to efficiently capture both asynchronous temporal dependencies and complex variable interdependencies. By decoupling the spatio-temporal relations into temporal and spatial components, STRGCN can handle a broad range of time series scenarios, including those with irregular timestamps and variable interactions. Moreover, the use of low-rank decompositions ensures that the model remains computationally efficient, even with a large number of variables, maintaining acceptable memory requirements.
For global application, we define a single update of the STRGCN layer on a graph as follows:

\begin{equation}
H^{(k+1)} 
= 
\mathrm{STRGCNLayer}\Bigl(H^{(k)}, P, S, \Theta^{(k)}\Bigr) 
\end{equation}
where \(H^{(k)}\) denotes the node embeddings at iteration \(k\). The parameters of the STRGCN layer are collectively denoted by \(\Theta\), defined as: 
\(
\Theta
=
\bigl\{
Q,\;\phi(\cdot),\;\{e_r\}_{r\in \mathcal{R}^S},\;W_0
\bigr\}.\)
\subsection{Hierarchical Sandwich Structure}
\begin{figure}[t!]
\centering
\includegraphics[width=0.9\columnwidth]{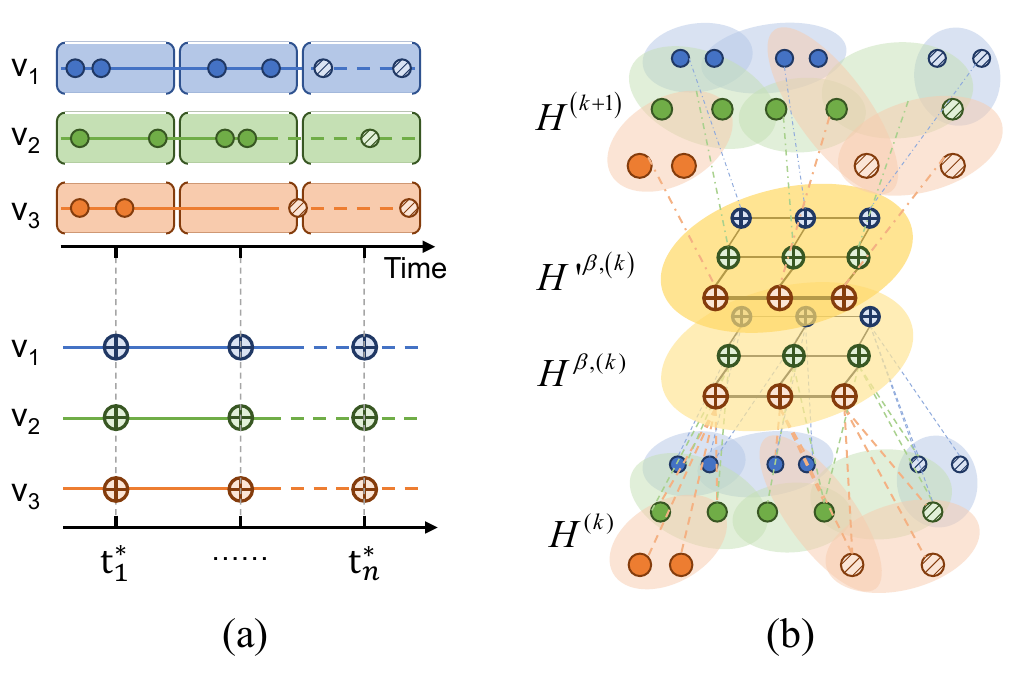}
\caption{Illustration of the Hyper-nodes Generation and the Hierarchical Sandwich Structure. Panel (a) shows the generation of hyper-nodes by uniformly sampling nodes along the temporal axis for each variable with a predefined window length. Panel (b) provides an intuitive understanding of the bottom, middle, and top layers in the Sandwich structure.}

\label{sandwich}
\end{figure}

In time series forecasting, individual observations often contain limited semantic information. Aggregating information from neighboring time points can yield more valuable local features \cite{nie2023a}. Furthermore, as the number of observations in an irregular multivariate time series (IMTS) increases, performing full graph convolutions on a fully connected graph becomes computationally prohibitive. To address this, we propose a hierarchical ``Sandwich'' structure that reduces computational costs while preserving essential spatio-temporal dependencies.

Specifically, we introduce the concept of a \emph{hyper-node}. A hyper-node corresponds to a group of nodes from the original graph, as illustrated in Figure \ref{sandwich}a. Hyper-nodes are generated by uniformly sampling nodes along the temporal axis with a predefined window length \(w\). The set of hyper-nodes is denoted as \(\mathcal{N}^{\beta}\). Let \(P^{\beta}\) and \(S^{\beta}\) denote the temporal encodings, and the variable ID for the hyper-nodes, respectively.

As shown in Figure \ref{sandwich}b,
From bottom to top, the sandwich structure is divided into four levels of representation, with a STRGCN applied between every two levels, which we respectively name bottom, middle, and top.
Initially, we use the STRGCN to enable each hyper-node in \(\mathcal{N}^{\beta}\) to perform information transfer with all lower-level nodes \(H^{(k)}\) to obtain the hidden states of hyper-nodes \(H^{\beta,(k)}\):
\begin{equation}
H^{\beta,(k)}
= 
\mathrm{STRGCNLayer}\Bigl(H^{(k)}, P, S, \Theta^{bottom,(k)}\Bigr)
\end{equation}
Unlike strict patch-based methods, each hyper-node adaptively aggregates information from its neighboring low-level nodes using a STRGCN layer, rather than rigidly corresponding to a fixed time interval.

The middle layer further aggregates representations at the level of the hyper-nodes:

\begin{equation}
H'^{\beta,(k)}
= 
\mathrm{STRGCNLayer}\Bigl(H^{\beta,(k)}, P^{\beta}, S^{\beta}, \Theta^{middle,(k)}\Bigr)
\end{equation}

We then upscale from hyper-nodes back to the original node resolution at the top layer:

\begin{equation}
H'^{(k+1)}
=
\mathrm{STRGCNLayer}\Bigl(H'^{\beta,(k)}, P^{\beta}, S^{\beta}, \Theta^{top,(k)}\Bigr)
\end{equation}

where \(\Theta^{bottom,(k)}\), \(\Theta^{middle,(k)}\), and \(\Theta^{top,(k)}\) are the parameters for each STRGCN layer at iteration \(k\).

Finally, a residual connection adds the top-layer representation \(H'^{(k+1)}\) to the original bottom-layer representation \(H^{(k)}\):

\begin{equation}
H^{(k+1)} 
= 
H'^{(k+1)} 
\;+\; 
H^{(k)}
\end{equation}


By introducing hyper-nodes and employing this hierarchical ``Sandwich" structure, the STRGCN reduces computational complexity through adaptive node aggregation, while effectively capturing both local and global spatio-temporal dependencies in irregular multivariate time series data.

\section{Experiments}
\subsection{Experimental Details}
\subsubsection{Datasets}
We evaluate the performance of our proposed model on four widely-recognized irregular multivariate time series (IMTS) datasets, representing diverse application domains. These datasets include two healthcare datasets, PhysioNet and MIMIC, a human activity recognition dataset, Activity, and a climate dataset, USHCN. To ensure a robust evaluation, we randomly partition each dataset into training, validation, and test sets using a 60\%:20\%:20\% split.

\subsubsection{Baselines} 

We select 18 state-of-the-art models across different domains as baselines. Specifically, we include IMTS forecasting models Latent ODEs \cite{rubanova2019latent}, mTAND \cite{shukla2021multitime}, CRU \cite{schirmer2022modeling}, Neural Flows \cite{bilovs2021neural}, and t-PatchGNN \cite{zhang2024irregular}. Additionally, we adapt several advanced MTS models for IMTS forecasting by employing interpolation strategies to handle irregular time series inputs, including DLinear \cite{zeng2023transformers}, TimesNet \cite{wu2023timesnet}, PatchTST \cite{nie2023a}, Crossformer \cite{zhang2023crossformer}, GraphWaveNet \cite{wu2019graph}, MTGNN \cite{wu2020connecting}, StemGNN \cite{cao2020spectral}, CrossGNN \cite{huang2023crossgnn}, and FourierGNN \cite{yi2024fouriergnn}. Furthermore, we also adapt models originally designed for IMTS classification, such as GRU-D \cite{che2018recurrent}, SeFT \cite{horn2020set}, RainDrop \cite{zhang2022graphguided}, and Warpformer \cite{zhang2023warpformer}, by replacing their classification heads with forecasting heads to perform the comparison.

\subsubsection{Setup}
All experiments were conducted on an NVIDIA GeForce RTX 4090 24GB GPU. To ensure a fair comparison, we adopted a consistent experimental setup identical to that of t-PatchGNN, and thus some of the results in Table \ref{Tab1} are directly sourced from the t-PatchGNN implementation. The Mean Squared Error (MSE) was employed as the loss function throughout the experiments. Specifically, for the PhysioNet and MIMIC datasets, we used 24 hours of historical data to forecast the next 24 hours, whereas the Activity dataset utilized 3000 ms of historical data to predict the subsequent 1000 ms. In the case of the USHCN dataset, 24 months of historical data were used to predict the next 1 month.

For more detailed information on the datasets, model descriptions, hyperparameter configurations, and other implementation details, please refer to the Appendix.

\subsection{Main Results}

Table \ref{Tab1} presents the predictive performance of STRGCN across four irregular multivariate time series forecasting datasets, demonstrating superior accuracy over current state-of-the-art models in most cases. Specifically, STRGCN achieved the best performance in terms of mean squared error (MSE) on all four datasets, and in terms of mean absolute error (MAE), STRGCN achieved optimal results on three datasets, with slightly lower performance on the Activity dataset, where it was outperformed by t-PatchGNN and Warpformer.

We attribute STRGCN's slightly weaker performance on the Activity dataset to the fact that the Activity dataset has a very small sampling interval and is largely regularly sampled, with missing values. STRGCN’s advantage is constrained by the complexity of its design for modeling asynchronous spatio-temporal relationships. For datasets like Activity, which exhibit strong regularity and nearly consistent temporal intervals with high resolution, the asynchronous spatio-temporal mechanism is overly redundant and may not effectively improve model performance.

Despite these challenges, STRGCN achieved an average MSE reduction of 9.89\% across four datasets compared to t-PatchGNN, the current state-of-the-art model. We believe that the patch-based temporal segmentation approach used in t-PatchGNN may lead to the loss of finer spatio-temporal semantics, while STRGCN improves performance by capturing asynchronous temporal relationships at both the node and local granularities. Additionally, we observe that the performance improvement of STRGCN is more significant in terms of MSE compared to MAE. This is likely because MSE is more sensitive to outliers, whereas MAE penalizes outliers less severely. Given that the graph convolutional network structure inherently applies a smoothing effect, it tends to suppress extreme fluctuations or outliers in the input data, thus leading to a greater reduction in MSE.

\subsection{Ablation Study}
To assess the effectiveness of the STRGCN module design, we conducted ablation studies on four datasets, evaluating the following variants: 
\textbf{w/o-TimeRel}: Removes the temporal relation component (\(W_r^t\) in Equation \ref{deST}), omitting temporal dependencies.
\textbf{w/o-VarRel}: Removes the spatial relation component (\(W_r^s\) in Equation \ref{deST}), excluding inter-variable dependencies.
\textbf{re-GCN}: Replaces STRGCN with a standard GCN, encoding timestamp and node features directly into node representations.
\textbf{w/o-Sandwich}: Removes the hierarchical sandwich structure.

\begin{table}[h!]
\centering
\renewcommand{\arraystretch}{1.3}
\resizebox{0.45\textwidth}{!}{\begin{tabular}{c|c|c|c|c|c}
\toprule
\multicolumn{2}{c|}{Dataset} & PhysioNet & MIMIC & Activity& USHCN \\
\cmidrule(lr){1-2} \cmidrule(lr){3-6}
\multicolumn{2}{c|}{Metric} & MSE\(\times10^{-3}\) & MSE\(\times10^{-2}\) & MSE\(\times10^{-3}\) & MSE\(\times10^{-1}\) \\
\midrule
\multicolumn{2}{c|}{STRGCN} & \textbf{3.98 ± 0.16} & \textbf{1.47 ± 0.05} & \textbf{2.53 ± 0.18} & \textbf{4.92 ± 0.06} \\
\multicolumn{2}{c|}{w/o-TimeRel} & 4.08 ± 0.12 & 1.50 ± 0.03 & 2.59 ± 0.09&  5.06 ± 0.04\\
\multicolumn{2}{c|}{w/o-VarRel} & 5.12 ± 0.09 & 1.70 ± 0.10 & 3.03 ± 0.17 & 5.18 ± 0.06 \\
\multicolumn{2}{c|}{re-GCN} & 5.47 ± 0.28 & 1.76 ± 0.04 & 3.86 ± 0.23 & 5.22 ± 0.03 \\
\multicolumn{2}{c|}{w/o-Sandwich} & 4.10 ± 0.11 & 1.55 ± 0.08 & - & 5.02 ± 0.07 \\

\bottomrule
\end{tabular}}
\caption{Ablation analysis of the STRGCN model, The best results are highlighted in \textbf{bold}. Note that since the hierarchical sandwich structure is not enabled by default in STRGCN for the Activity dataset, the results for w/o-Sandwich are omitted.}
\label{Tab2}
\end{table}

Table \ref{Tab2} shows the ablation results. Removing temporal dependencies (i.e., \textbf{w/o-TimeRel}) leads to a performance drop, highlighting the importance of modeling asynchronous temporal relations. Excluding spatial dependencies (i.e., \textbf{w/o-VarRel}) results in a 16.8\% average increase in MSE, emphasizing the role of inter-variable relationships in multivariate time series forecasting.

Replacing STRGCN with a standard GCN (i.e., \textbf{re-GCN}) results in significantly worse performance, suggesting that explicitly modeling spatio-temporal dependencies is more effective than encoding them in node features alone. Finally, removing the sandwich structure (i.e., \textbf{w/o-Sandwich}) increases error and memory usage, as discussed in the next section.

\subsection{Model Efficiency}

\begin{figure}[h!]
\centering
\includegraphics[width=1\columnwidth]{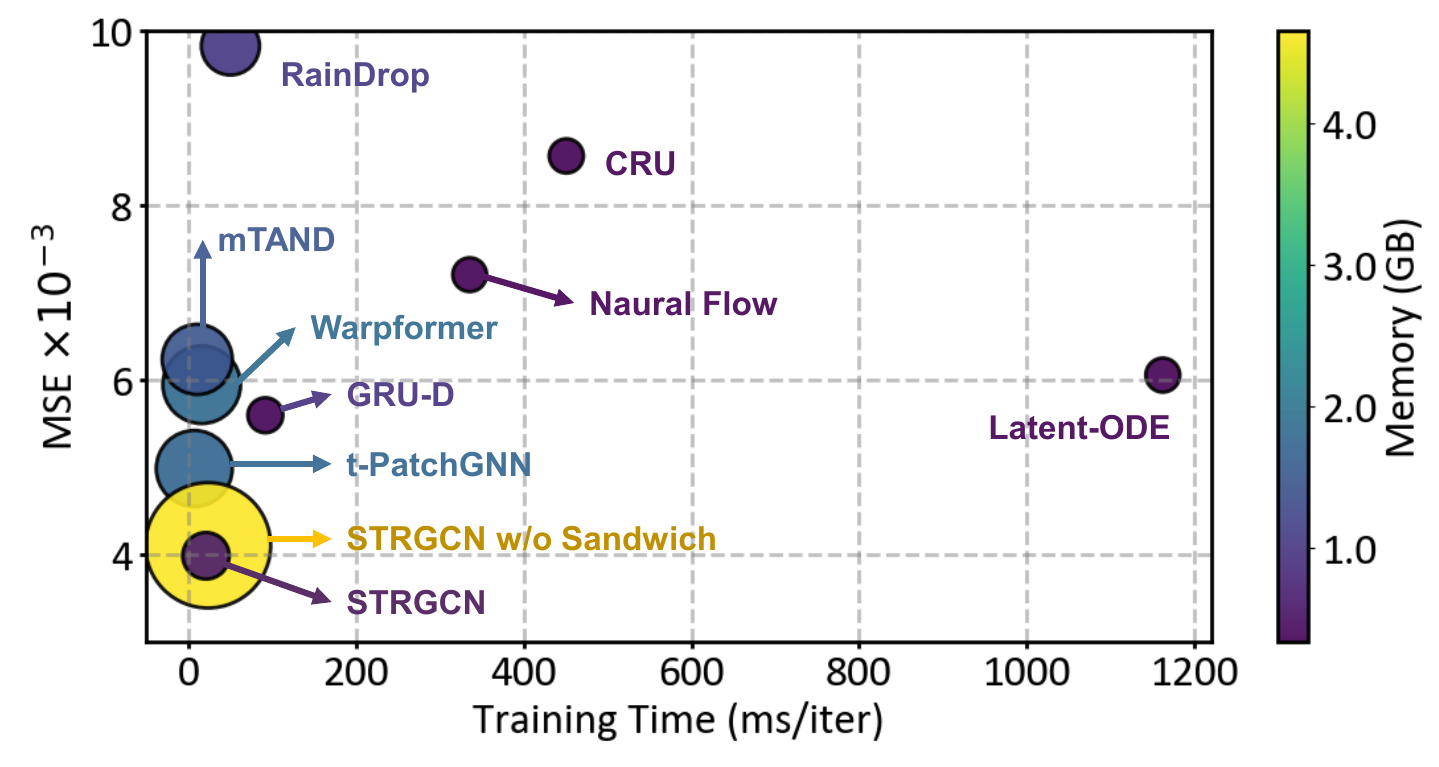} 
\caption{Performance Analysis of STRGCN: Assessing Average MSE, Training Time, and Memory Usage (lower is better), evaluated on the PhysioNet Dataset.}
\label{fig4}
\end{figure}

We compare the STRGCN model with other state-of-the-art models across three key aspects: forecasting accuracy, memory usage, and training speed. As shown in Figure \ref{fig4}, STRGCN outperforms the comparison models in predictive accuracy. Concurrently, it displays the best overall performance with relatively lower memory consumption and faster training speed. 

Notably, the inclusion of the hierarchical sandwich structure in STRGCN results in a significant reduction in memory usage, achieving a 720\% decrease compared to the model without this structure.

\bibliographystyle{named}
\bibliography{ijcai25}

\end{document}